\documentclass{article}

\usepackage[accepted]{icml2023}
\usepackage{microtype}
\usepackage{graphicx}
\usepackage{subfigure}
\usepackage{booktabs} 

\usepackage{hyperref}

\usepackage{icml2023}

\usepackage{amsmath}
\usepackage{amssymb}
\usepackage{mathtools}
\usepackage{amsthm}

\usepackage[capitalize,noabbrev]{cleveref}

\theoremstyle{plain}

\theoremstyle{definition}

\theoremstyle{remark}

\usepackage[textsize=tiny]{todonotes}

\icmltitlerunning{Submission and Formatting Instructions for ICML 2023}

\begin{document}

\twocolumn[
\icmltitle{Transcending Traditional Boundaries: Leveraging Inter-Annotator Agreement (IAA) for Enhancing Data Management Operations (DMOps)}




\begin{icmlauthorlist}
\icmlauthor{Damrin Kim}{yyyd,yyyc}
\icmlauthor{NamHyeok Kim}{yyyc}
\icmlauthor{Chanjun Park}{yyyc}
\icmlauthor{Harksoo Kim}{yyyd}
\end{icmlauthorlist}
  
\icmlaffiliation{yyyc}{Upstage, Gyeonggi-do, Korea}
\icmlaffiliation{yyyd}{Konkuk University, Seoul, Korea}


\icmlcorrespondingauthor{Chanjun Park}{chanjun.park@upstage.ai}
\icmlcorrespondingauthor{Harksoo Kim}{nlpdrkim@konkuk.ac.kr}

\icmlkeywords{Machine Learning, ICML}

\vskip 0.3in
]



\printAffiliationsAndNotice{}  

\begin{abstract}
This paper presents a novel approach of leveraging Inter-Annotator Agreement (IAA), traditionally used for assessing labeling consistency, to optimize Data Management Operations (DMOps). We advocate for the use of IAA in predicting the labeling quality of individual annotators, leading to cost and time efficiency in data production. Additionally, our work highlights the potential of IAA in forecasting document difficulty, thereby boosting the data construction process's overall efficiency. This research underscores IAA's broader application potential in data-driven research optimization and holds significant implications for large-scale data projects prioritizing efficiency, cost reduction, and high-quality data. 
\end{abstract}

\section{Introduction}
Within industry settings, data specificity to principal business domains is crucial, particularly for B2B companies requiring data congruent with customer needs and business objectivest~\cite{zha2023data}. This has led many companies to create proprietary data and stimulated research into efficient data construction pipelines, namely Data Management Operations (DMOps)~\cite{choi2023dmops}. 

In the pursuit of time and cost efficiency in DMOps, preempting labeling errors to minimize data quality degradation is of paramount importance. This necessitates effective forecasting of data quality, data production time and cost, and judicious task allocation amongst workers. Consequently, the ability to predict data quality in advance to curb unnecessary costs and forestall worker errors is vital to enhancing data quality~\cite{ge2017predicting}.

In this paper, we present an innovative application of the Inter Annotator Agreement (IAA)~\cite{artstein2017inter}, a metric traditionally used for assessing labeling consistency, to augment the efficiency of DMOps. We first demonstrate how IAA can be employed to predict individual workers' labeling quality. By measuring and comparing the IAA scores of workers, we identify those whose scores significantly deviate from the norm, visualized via a heatmap. This approach facilitates targeted retraining, preemptively mitigating quality disparities amongst workers and naturally leading to improved data quality, time, and cost reductions.

Further, we illustrate how IAA can be leveraged to predict document complexity. By gauging the IAA score of documents during the pilot phase and comparing these with the scores of pre-existing documents, we can preemptively forecast document difficulty. This capability enables effective planning for modeling efforts and task allocation, bolstering the overall operation's efficiency.

However, this paper aims to apply IAA not only to measure the level of agreement in data but also to predict data quality and improve DMOps process efficiency.



\section{Predictive Analysis of Labeling Quality by Worker}
\paragraph{Task} In this study, we have chosen to focus on the task of Visual Information Extraction (VIE)~\cite{9412927} underpinned by Optical Character Recognition (OCR) technology. The complexity of this task necessitates the labeling of key-value parsing information, enabling the extraction of relevant data from the recognized text, the understanding of structured information, and facilitating subsequent analyses or processing. 

\paragraph{Setting} Our experimental design incorporates a variety of official documents. This includes certificate of resident registration (10 pages), certificate of archived family relations registration (10 pages), certificate of family relations (10 pages), and certificate of basic residence registration (10 pages). These documents collectively encompass a wide range of information, including personal details, family relationships, and specific identification data such as birth dates, genders, and issuance dates. This information necessitates tagging by the assigned workers. Our experimental procedure engaged three workers (A, B, C) for this task.

\begin{figure}[!htbp]
\centering
\includegraphics[height = 5.5cm]{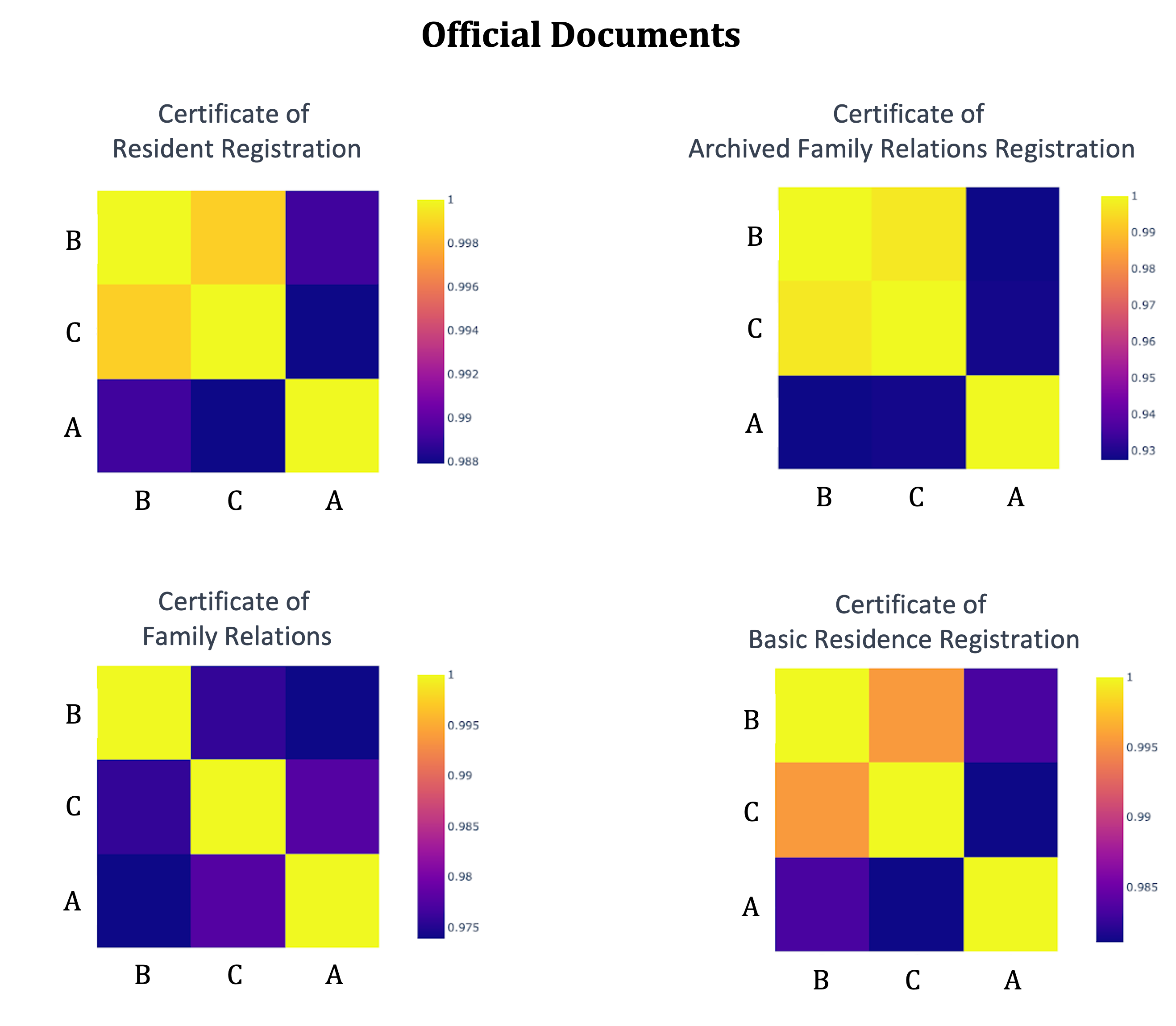}
\caption{Kappa coefficients for each worker in the four official documents}
\hfil
\label{fig:fig1}
\end{figure}

\paragraph{Results} The quality prediction of labeling per worker was examined through the experimental results delineated in Figure \ref{fig:fig1}. Our primary objective was to visually render the Inter-Annotator Agreement (IAA) results in a heatmap format to facilitate an intuitive understanding of discrepancies among various workers.

As per the heatmap analysis illustrated in Figure \ref{fig:fig1}, there are observable discrepancies between worker A and workers B \& C with respect to certificate of resident registration and certificate of archived family relations registration. This observation lends itself to the inference that additional training and intervention might be requisite for worker A to enhance the quality of the assigned tasks. Specifically, if the IAA outcomes are numerically deficient or if significant discrepancies surface among workers, further work might be necessitated to bolster the reliability and precision of data quality. This insight could aid in optimizing the efficiency of the DMOps process by leveraging the IAA.

Conversely, the Kappa coefficient for workers A, B, and C concerning certificate of family relations and certificate of basic residence registration exhibits an exceptionally high average of 0.97. This implies that these datasets are of a very high quality. Hence, we infer that these two document types may not necessitate any additional work.

In conclusion, IAA serves as a valuable tool for measuring the labeling quality per worker. If significant inconsistencies among workers are detected from these measurements, it may be prudent to identify the worker(s) in question and consider rework. Instead, it should be reserved for situations where quantitative measurements exhibit substantial discrepancies or fall below a pre-determined threshold. By adopting this strategic approach to data quality prediction, we can effectively reduce the time and cost associated with the work process, leading to a more efficient and streamlined construction process.




\section{Analyzing Document Difficulty Prediction}
For our second experiment, we compared the IAA between various documents, including official documents, to determine whether IAA could be used as a predictor of document difficulty. If document difficulty can be predicted, it is possible to plan the modeling effort and worker allocation in advance, thus improving the overall efficiency of the task.

The experiment results are shown in Table \ref{tab:t1}, with IAA values obtained in the following order: official documents, diagnoses and medical reports, requests for quotes, and bills of lading. Generally, documents that are more complex and contain more parsing information tend to be more difficult to work with, and the same trend was observed in this experiment.
These results suggest that IAA can not only be used as a measure of consistency but also as a useful indicator for measuring the difficulty of documents.

\begin{table}[]
\caption{Comparison and analysis table of IAA results by document}
\label{tab:t1}
\LARGE
\resizebox{\columnwidth}{!}{%
\begin{tabular}{ccccc}
\addlinespace
\hline
\addlinespace
\textbf{Metrics}                   & \textbf{\parbox{3.5cm}{Official\newline\centering Documents}} & \textbf{\parbox{4cm}{Diagnosis and\newline\centering Medical Report}} & \textbf{\parbox{3.5cm}{Request\newline\centering for Quote}} & \textbf{\parbox{3.5cm}{Bill\newline\centering of Lading}} \\ 
\addlinespace
\hline
\addlinespace
\textbf{Cohen's Kappa}                   & 0.976          & 0.967           & 0.927          & 0.772         \\
\addlinespace
\textbf{Fleiss's Kappa}                  & 0.976          & 0.967           & 0.927          & 0.771         \\
\addlinespace
\textbf{Krippendorff's Alpha}            & 0.978          & 0.969           & 0.933          & 0.778         \\ \hline
\addlinespace
\end{tabular}%
}
\end{table}

\section{Conclusions}
This paper conducted research to increase the efficiency of DMOps using Inter Annotator Agreement (IAA). The study demonstrated that IAA can be used for predicting labeling quality by worker, measuring document difficulty. This makes it possible to anticipate the difficulty of documents in advance and plan for modeling efforts and worker allocation accordingly. In other words, IAA can be used not only as a measure of labeling consistency, but also as a useful indicator for overall data construction process improvement.

\section*{Acknowledgements}
This work was supported by Institute for Information \& communications Technology Planning \& Evaluation(IITP) grant funded by the Korea government(MSIT) (No. 2022-0-00369, (Part 4) Development of AI Technology to support Expert Decision-making that can Explain the Reasons/Grounds for Judgment Results based on Expert Knowledge)

\nocite{langley00}

\bibliography{example_paper}

\begin{thebibliography}{5}
\providecommand{\natexlab}[1]{#1}
\providecommand{\url}[1]{\texttt{#1}}
\expandafter\ifx\csname urlstyle\endcsname\relax
  \providecommand{\doi}[1]{doi: #1}\else
  \providecommand{\doi}{doi: \begingroup \urlstyle{rm}\Url}\fi

\bibitem[Artstein(2017)]{artstein2017inter}
Artstein, R.
\newblock Inter-annotator agreement.
\newblock \emph{Handbook of linguistic annotation}, pp.\  297--313, 2017.

\bibitem[Choi \& Park(2023)Choi and Park]{choi2023dmops}
Choi, E. and Park, C.
\newblock Dmops: Data management operation and recipes.
\newblock \emph{arXiv preprint arXiv:2301.01228}, 2023.

\bibitem[Ge et~al.(2017)Ge, O’Brien, and Helfert]{ge2017predicting}
Ge, M., O’Brien, T., and Helfert, M.
\newblock Predicting data quality success-the bullwhip effect in data quality.
\newblock In \emph{Perspectives in Business Informatics Research: 16th
  International Conference, BIR 2017, Copenhagen, Denmark, August 28--30, 2017,
  Proceedings 16}, pp.\  157--165. Springer, 2017.

\bibitem[Yu et~al.(2021)Yu, Lu, Qi, Gong, and Xiao]{9412927}
Yu, W., Lu, N., Qi, X., Gong, P., and Xiao, R.
\newblock Pick: Processing key information extraction from documents using
  improved graph learning-convolutional networks.
\newblock In \emph{2020 25th International Conference on Pattern Recognition
  (ICPR)}, pp.\  4363--4370, 2021.
\newblock \doi{10.1109/ICPR48806.2021.9412927}.

\bibitem[Zha et~al.(2023)Zha, Bhat, Lai, Yang, Jiang, Zhong, and
  Hu]{zha2023data}
Zha, D., Bhat, Z.~P., Lai, K.-H., Yang, F., Jiang, Z., Zhong, S., and Hu, X.
\newblock Data-centric artificial intelligence: A survey.
\newblock \emph{arXiv preprint arXiv:2303.10158}, 2023.

\end{thebibliography}
\bibliographystyle{icml2023}

\end{document}